\documentclass[10pt,twocolumn,letterpaper]{article}

\usepackage{cvpr}
\usepackage{times}
\usepackage{epsfig}
\usepackage{graphicx}
\usepackage{amsmath}
\usepackage{amssymb}
\makeatletter
\@namedef{ver@everyshi.sty}{}
\makeatother
\usepackage{tikz}
\usepackage{bbm}
\usepackage{pgfplots}
\usepgfplotslibrary{colorbrewer}
\pgfplotsset{cycle list/Set2}
\usepackage{subfig}
 \pgfplotsset{
        cycle list/Set2,
        cycle multiindex* list={
            mark list*\nextlist
            Set2\nextlist
        },
    }
\usepgfplotslibrary{groupplots}
\usetikzlibrary{patterns}

\usepackage[breaklinks=true,bookmarks=false]{hyperref}

\cvprfinalcopy 

\def\cvprPaperID{****} 

\ifcvprfinal\fi
\begin{document}

\title{Deep Spherical Quantization for Image Search}

\author{Sepehr Eghbali and Ladan Tahvildari\\
University of Waterloo\\
{\tt\small \{sepehr.eghbali,ladan.tahvildari\}@uwaterloo.ca}
}

\maketitle
\thispagestyle{empty}

\begin{abstract}
%

Hashing methods, which encode high-dimensional images with compact discrete codes, have been widely applied to enhance large-scale image retrieval. In this paper, we put forward Deep Spherical Quantization (DSQ), a novel method to make deep convolutional neural networks generate supervised and compact binary codes for efficient image search. Our approach simultaneously learns a mapping that transforms the input images into a low-dimensional discriminative space, and quantizes the transformed data points using multi-codebook quantization. To eliminate the negative effect of norm variance on codebook learning, we force the network to $L_2$ normalize the extracted features and then quantize the resulting vectors using a new supervised quantization technique specifically designed for points lying on a unit hypersphere. Furthermore, we introduce an easy-to-implement extension of our quantization technique that enforces sparsity on the codebooks. Extensive experiments demonstrate that DSQ and its sparse variant can generate semantically separable compact binary codes outperforming many state-of-the-art image retrieval methods on three benchmarks.
\end{abstract}

\section{Introduction}

Nearest neighbor search is one of the fundamental problems in multimedia systems. Given a query point, the goal entails finding the most similar item to the query in a dataset. Accuracy and speed are two key aspects in retrieval systems, however, with explosive growth of high dimensional items such as images, videos and documents on the Internet, most of traditional branch and bound indexing data structures are deemed impractical, mainly because of their query time or memory cost that grow exponentially with the number of dimensions. This has led to a burgeoning field of research, \textit{Approximate Nearest Neighbor} (ANN), that focuses on reducing storage and computational costs with minimal accuracy loss.


ANN problem has witnessed a great amount of research over the past two decades. The state-of-the-art in ANN is mainly focused on \textit{hashing} (compact coding), which aims at encoding high-dimensional media data into short binary codes subject to preserving a given notation of similarity. Binary-valued representation has several advantages, such as being compact to store and faster to compare, making it a suitable fit for large-scale nearest neighbor search. Moreover, for binary strings, one can achieve sublinear query time using hash tables~\cite{eghbali2019fast,norouzi2014fast} or tree-based indexing data structures~\cite{eghbali2017online,eghbali2019online}. Finding compact binary codes that better respect the given notion of similarity has been the topic of much work over the last two decades during which a rich set of hashing techniques has been proposed.
Compact coding techniques are roughly in two streams, categorized by the way they compute distance between encoded items: 1) \textit{Binary hashing} maps high-dimensional input vectors into Hamming space where the distance between two codes can be computed extremely fast using bitwise operators. 2) \textit{Multi-Codebook Quantization} (MCQ) which, analogous to k-means algorithm, partitions the input space into non-overlapping cells and then approximates the distance between two points with the distance between the centers of cell they belong to. The search speed enhancement of MCQ stems from the fact that the distance between cells can be pre-computed and stored in lookup tables.

Not surprisingly, with the dawn of deep learning, most of recent research effort in compact coding has been directed towards using deep networks for producing compact and functional binary codes. Deep hashing methods simultaneously learn the representation and hash coding from raw images. Similarly, deep MCQ has been the topic of study in recent years~\cite{cao2016deep,jain2017subic}. Surprisingly enough, although MCQ is a more powerful model as it enables producing many more possible distinct distances, due to the lack of research, its performance in the context of deep supervised compact coding is inferior to state-of-the-art in supervised binary hashing~\cite{li2017deep}. 

Most of existing deep supervised MCQ techniques incorporate an unsupervised quantization (usually product quantization (PQ)~\cite{jegou2011product}) on top of the features generated by a deep architecture. Nevertheless, the adopted networks often produce deep features with relatively high norm variance which adversely affects the quality of quantization~\cite{wu2017multiscale}. To address this shortcoming, we reformulate the quantization problem by $L_2$ normalizing deep features to remove norm variance. By exploiting the fact that resulting features lie on a hypersphere, we propose a novel MCQ algorithm that drops the hard orthogonality constraint of product quantization to achieve lower quantization error.  Furthemore, to encourage better discriminating performance, inspired by the recently proposed center loss~\cite{wen2016discriminative}, we add a supervised quantization loss term to the final objective function to increase inter-class variance.  Finaly, we propose a sparse extension of our quantization algorithm which is necessary for dealing with large codebooks~\cite{zhang2015sparse}. Comprehensive empirical studies on three standard image retrieval benchmarks testify that DSQ generates compact binary codes which outperform many state-of-the-art methods.






\section{Related Work}

Existing hashing methods consist of supervised and unsupervised hashing. We refer interested readers to~\cite{wang2018survey} for a comprehensive survey. 

Unsupervised hashing methods learn hash functions that map data to binary codes using unlabeled data. Typical learning criteria are reconstruction error minimization~\cite{kulis2009learning}, preserving local neighborhood~\cite{liu2014discrete} and quantization error minimization~\cite{gong2013iterative}. Supervised hashing, on the other hand, aims at learning binary codes that are faithful to a given notion of semantic information such as point-wise (class labels)~\cite{li2017deep,shen2015supervised,wang2016supervised}, pairwise~\cite{cao2018hashgan,Cao_2018_CVPR,Chen_2018_CVPR} or triplet labels~\cite{liu2018deep,norouzi2012hamming}.

\textbf{Multi-Codebook Quantization.}
A subclass of unsupervised hashing methods, called Multi-Codebook Quantization (MCQ), is formulated as a quantization problem which aims at approximating vectors with summation of multiple codewords. Formally, let $X=[\mathbf{x}_1, \ldots, \mathbf{x}_n] \in \mathbb{R}^{d \times n}$ denote the set of $n$ points to be quantized, MCQ is the problem of finding 1) $m$ codebooks  (dictionaries) $C_j \in \mathbb{R}^{d \times h}, j \in \{1\, \ldots,m\}$ each containing $h$ codewords, and 2) the  encoding binary vectors $\mathbf{b}_i = [\mathbf{b}_{i1}^T, \ldots, \mathbf{b}_{im}^T]^T \in \{0,1\}^{mh\times 1}$, that minimize the quantization error:

\begin{equation}
\label{eq::MCQ}
\sum_{i=1}^n \| \mathbf{x}_i - [C_1,\ldots,C_m]\mathbf{b}_i\|_2^2
\end{equation}
where each subcode $\mathbf{b}_{ij}$ is limited to having only one non-zero entry, $\|\mathbf{b}_{ij}\|_1=1$, to ensure only one codeword per codebook is selected. During the query phase, MCQ uses the approximation of each point to estimate the distance between the query $\mathbf{q}\in\mathbb{R}^d$ and each data point:
\begin{equation}
\begin{split}
\|\mathbf{q} - \mathbf{x}_i\|_2^2 & \approx \sum_{i=1}^m \|\mathbf{q} - C\mathbf{b}_i\|_2^2 - (m-1) \|\mathbf{q}\|_2^2 \\
& +
 \sum_{t\neq j} (C_t\mathbf{b}_{it})^T C_j\mathbf{b}_{ij}
\end{split}
\label{eq::distibutive}
\end{equation}

Given the query, the first term can be efficiently computed using lookup tables that store the distance between the query and each codeword. The second term can be ignored during search time as it is constant for a given fixed query. One of the key features that differentiates MCQ techniques is how they handle the third term. Product quantization (PQ)~\cite{jegou2011product}, Cartesian K-means (CKM)~\cite{norouzi2013cartesian} and Optimized Product Quantization (OPQ)~\cite{ge2014optimized} restrict the codebooks to be mutually orthogonal making the third term equal to zero. Composite Quantization~\cite{wang2018composite}, on the other hand, forces it to be a constant which in turn makes the resulting optimization problem hard to solve. 

Additive quantization (AQ)~\cite{babenko2014additive} and its enhanced extension Local Search Quantization (LSQ)~\cite{martinez2016revisiting} expand $\|\mathbf{q} - \mathbf{x}_i\|_2^2$ based on the inner product of query and codewords but their formulation requires not only approximating the input vector but also its $L_2$ norm $\|\mathbf{x}_i\|_2^2$. AQ provides two solutions to estimate the norm. The first is to separately quantize the scalar value $\|\mathbf{x}_i\|_2^2$, which results in an additional memory cost that grows linearly with the database size. The other way is to estimate the norm with the codewords
which makes the cost of distance computation quadratic in the number of codebooks.

\textbf{Supervised MCQ.}
While most of the research in supervised hashing is focused on supervised binary hashing, a handful of studies have been recently proposed on using MCQ in the supervised setting. Supervised MCQ techniques can be, for the most part, described as a combination of supervised loss function and one of the unsupervised MCQ techniques described above. Supervised Quantization (SQ)~\cite{wang2016supervised} combines supervised $L_2$ loss with CQ, however, the resulting optimization problem is hard to solve as it inherits the constant inter-dictionary-element-product constraint from CQ. Deep Quantization Network (DQN)~\cite{cao2016deep} combines a deep architecture and PQ. One shortcoming of DQN is that during codebook optimization, it ignores the supervisory information. SUBIC~\cite{jain2017subic} integrates the one hot-encoding layer in deep neural network which encodes each image with concatenation of one-hot block similar to PQ. However, its sparse property limits its representation capability.

\begin{figure}
\centering
\begin{tikzpicture}
\begin{axis}[
    ybar,
    height=5cm,
    width=7cm,
    xmajorgrids,
	ymajorgrids,
   legend pos=north west,
                       label style={font=\small},
                    tick label style={font=\small},  
   legend style={font=\fontsize{7}{5}\selectfont},
    ylabel={MAP},
    symbolic x coords={DQN~\cite{cao2016deep},DTQ~\cite{liu2018deep}},
    xtick=data,
    ]
\addplot+[mark=none,postaction={
        pattern=north east lines
    }] coordinates {(DQN~\cite{cao2016deep},0.6133) (DTQ~\cite{liu2018deep},0.7373)};
\addplot+[mark=none,postaction={
        pattern=north east lines
    }] coordinates {(DQN~\cite{cao2016deep},0.6189) (DTQ~\cite{liu2018deep},0.7412)};
\legend{Unnormalized,Normalized}
\end{axis}
\end{tikzpicture}
\caption{Performance of two supervised MCQ models with and without feature normalization on CIFAR-10 for 64-bit codes.}
\label{fig::normalization}
\end{figure}
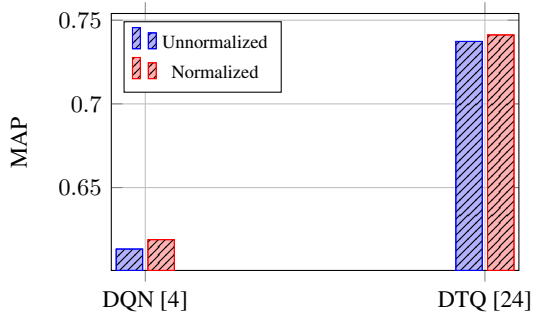

\textbf{Effect of Norm Variance on MCQ.}
Recently, Wu \etal ~\cite{wu2017multiscale} have shown that norm variance adversely affects the quantization error of unsupervised MCQ techniques, even when the variance is relatively moderate. To address this issue, authors propose to separately scalar quantize the data point norms and then unit normalize the data points before applying PQ. Nevertheless, it is not clear how the quantization budget should be split between the two the quantizers. Also, PQ imposes strong orthogonality on the codebooks which reduces the fidelity of leaned codebooks~\cite{babenko2014additive}. 

We conclude this section by empirically showing the effect of norm variance on the performance of supervised quantization. To this aim, we run two state-of-the-art supervised MCQ techniques with and without feature normalization on CIFAR-10 dataset (settings of the experiment will be discussed in Section~\ref{sec::experiments}). Figure~\ref{fig::normalization} plots the MAP performance of two supervised MCQ techniques and it demonstrates that one can achieve marginal performance gain by simply normalizing the features during training without incurring any additional cost.

\section{Proposed Approach}
In similarity retrieval, we are given a training set of $n$ points, $\mathcal{X}=\{\mathbf{x}_i \in \mathbb{R}^d\}_{i=1}^n$, with each point associated with a class label, $y_i \in  \{1,\ldots,l\}$. 
The goal, given query point $\mathbf{q}\in \mathbb{R}^d$, entails (approximately) finding items in $\mathcal{X}$ that  are semantically closest to $\mathbf{q}$ so that the found neighbors share the same class label as $\mathbf{q}$. This paper follows the idea of compact coding techniques that is converting database vectors into compact code and then performing the similarity search in the resulting space which has the advantage of lower memory cost and fast distance computation.

In this paper, we propose to use a deep network that maps the input points into a discriminative space, and simultaneously perform a form of a supervised MCQ on the embedded points to achieve fast retrieval with low computational and storage overhead. To this aim, we define a loss function comprising four terms, softmax loss, center loss, quantization loss, and discriminative loss each of which will be discussed in the following.

\subsection{Softmax and Center loss}

 In deep retrieval systems, obtaining a robust and discriminative representation is crucial for achieving good performance. Usually, this is achieved by applying the softmax loss to the representation layer of the network. However, the resulting features optimized with the supervision of softmax loss are often not discriminative enough as the softmax loss only focuses on finding a decision boundary that separates different classes without considering the intra-class compactness which is crucial to the accuracy of nearest neighbor search~\cite{he2018triplet,wen2016discriminative}.

To increase the intra-class variations while keeping the features of different classes separable, we adopt the state-of-the-art \emph{center loss}~\cite{wen2016discriminative} on top of the softmax loss.

Let $f(\cdot;\theta):\mathbb{R}^d \rightarrow \mathbb{R}^p$, with $p \ll d$, denote the feed-forward network that embed the input vectors into $p$-dimensional deep features, also let $\mathbf{z}_i$ denote deep feature representation of input $\mathbf{x}_i$, $\mathbf{z}_i=f(\mathbf{x}_i;\theta)$, then, the center loss is defined as:

\begin{equation}
\label{eq::center_loss}
    L_{C}= \sum_{i=1}^n\|\mathbf{z}_i-\phi_{y_i}\|_2^2
\end{equation}
where $y_i$ is the classes label associated with $\mathbf{z}_i$ and $\phi_{y_i}$ denotes the $y_i$-th class center of deep features.  Intuitively, center loss learns a center for the features of each class and meanwhile aims at pulling the deep features of the same class close to its corresponding center. It has been shown that joint supervision of softmax loss and center loss can produce significantly better discriminative deep features~\cite{wen2016discriminative}.

\subsection{Quantization loss}

We constrain the deep features to live on a $p$-dimensional unit hypersphere, \ie $\|f(\mathbf{x};\theta)\|_2=1$. Other than decreasing the intra-class variability of deep features~\cite{Wang_2018_CVPR}, there are two advantages in normalizing feature vectors:
1) norm variance is strictly zero, and 2) Euclidean nearest neighbor search is equivalent to Maximum Inner Product Search (MIPS) as for unit norm vectors we have $\|\mathbf{q} - \mathbf{x}\|_2^2 = 2 - 2\mathbf{q}^T\mathbf{x}$.

The main benefit of dealing with MIPS is that, unlike Euclidean distance (see~(\ref{eq::distibutive})), inner product naturally satisfies the distributive law, that is $\langle \mathbf{q},\sum_j \mathbf{t}_j\rangle= \sum_j \langle \mathbf{q},\mathbf{t}_j\rangle$. MCQ works well in large part due to the fact that it permits the distance between query and a quantized point to be computed as the summation of partial distances between query and selected codewords. Given the query, the distances between query and all codewords are stored in query-specific lookup tables and then used to calculate the distance between query and all quantized points. However, to make Euclidean distance satisfy the distributive law, we either need to enforce strong~\cite{jegou2011product,norouzi2013cartesian}/weak~\cite{zhang2014composite,zhang2015sparse} orthogonality constraints over the codewords of different dictionaries which reduces the fidelity of model and often leads to non-convex optimization, or we have to store the inner product between the all codewords in lookup table~\cite{babenko2014additive,martinez2016revisiting} which increases storage cost and distance computation time.

To reduce the approximation error of MIPS, we need to minimize the~\textit{distance reconstruction error} of MCQ. Since the Euclidean distance on the unit sphere is equal to the negative dot product plus a constant, distance reconstruction error can be rewritten as:


\begin{equation}
\begin{split}
    \mathbb{E}_{\mathbf{q}\sim P(\mathbf{q})} 
    &\Big[\sum_{i=1}^n|\langle\mathbf{z}_q,\mathbf{z}_i\rangle -
    \langle\mathbf{z}_q,\mathbf{\bar{z}}_i\rangle|
    \Big] = \\
    \mathbb{E}_{\mathbf{q}\sim P(\mathbf{q})} 
    &\Big[\sum_{i=1}^n \langle\mathbf{z}_q,\mathbf{z}_i
    - \mathbf{\bar{z}}_i\rangle 
    \Big] \leq \\
    & \sum_{i=1}^n\|\mathbf{z}_i - \mathbf{\bar{z}}_i\|_2
\end{split}
\end{equation}
where $\mathbf{\bar{z}}_i$ denotes the approximation of $\mathbf{z}_i$ using MCQ and $\mathbf{z}_q=f(\mathbf{q};\theta)$.

This suggests that the search accuracy directly depends on the quantization error; low quantization error leads to high search accuracy.


Therefore, the cost function we aim to optimize is the quantization loss:
\begin{equation}
\label{eq::spherical_quantization}
\begin{aligned}
    &L_{Q}(\{C_j\},\{\mathbf{b}_i\}) =  & & \sum_{i=1}^n\|\mathbf{z}_i - [C_1,\ldots,C_m]\mathbf{b}_i\|_2^2 \\
    & & & \mathbf{b}_i = [\mathbf{b}_{i1}^T,\ldots,\mathbf{b}_{im}^T]^T \\
    & & & \mathbf{b}_{ij} \in \{0,1\}^h, \|\mathbf{b}_{ij}\|_1=1 \\ 
    & & & j=1,\ldots m
\end{aligned}
\end{equation}
The benefit of such a simple formulation, in comparison to those that enforce multiple constraints on the codewords~\cite{jegou2011product,norouzi2013cartesian,zhang2014composite} are multi-fold; it causes a straightforward optimization procedure and also less implementation overhead.

\subsection{Discriminative Dictionary Learning}

Finally, we also incorporate the supervisory information during quantization procedure. In particular, we encourage the quantized points to be closer to their centers. To achieve this goal, we use the following loss:
\begin{equation}
\label{eq::disc}
    L_{D} = \sum_{i=1}^n \|\phi_{y_i} - C\mathbf{b}_i\|_2^2
\end{equation}
Intuitively, (\ref{eq::disc}) penalizes the cases where the point $\mathbf{\bar{z}}_i$ is not assigned to the clusters that are close to $\phi_{y_i}$.

The overall loss for training model takes the form:

\begin{equation}
\label{eq::final_optimization}
    L = L_{softmax} + \alpha L_Q + \lambda L_{C} + \gamma L_{D}
\end{equation}
where $\alpha, \lambda$ and $\gamma$ are the hyper-parameters that control the effect of each term.

\subsection{Optimization}
The objective function composes of four sets of learnable parameters, the parameters of the deep network $\theta$, the centers $\phi_{y_i}$s, the codewords in matrix $C$, the codeword assignment matrix $B$. We use alternative optimization to solve the problem with each iteration updating one set of parameters while fixing others. 


\textbf{Updating $\theta$}.
With $C$, $\phi_{y_i}$s, and $B$ fixed, the parameters of the network are updated through back-propagation as all of the terms in the loss are differentiable.

\textbf{Updating $\Phi$}.
We follow a similar procedure to~\cite{wen2016discriminative} for updating the centers. In particular, to avoid large perturbation caused by few mislabelled instances, we use a learning rate parameter $\zeta$ for training the centers:
\begin{equation}
\label{eq::update_c}
\mathbf{\phi}_{y_i}^{t+1} = \mathbf{\phi}_{y_i}^{t} - \zeta\Delta \mathbf{\phi}_{y_j}
\end{equation}
\begin{equation}
\Delta \mathbf{\phi}_{y_j}=\frac{\sum_{i=1}^n \mathbbm{1}(y_i=j)\cdot \left[\lambda(\phi_{y_i} -\mathbf{z}_i) + \gamma(\phi_{y_i} -C\mathbf{b}_i)\right]}
{1+\sum_{i=1}^n \mathbbm{1}(y_i=j)}
\end{equation}
where $\mathbbm{1}(condition)$ equals 1 if the condition is satisfied and 0 otherwise. Ideally, the centers should be updated in each iteration based on the whole training set which would be extremely costly. To reduce the cost, the update is performed on the mini-batches.

\textbf{Updating $C$}.
Given $B$, $\phi_{y_i}$s and $\theta$ fixed, the resulting optimization problem is:
\begin{equation}
\label{eq::codebook_problem}
    \alpha\|Z - CB\|_2^2 + \gamma\|\Phi - CB\|_2^2
\end{equation}
where $Z = [\mathbf{z}_1,\ldots,\mathbf{z}_n]$, $B=[\mathbf{b}_1,\ldots,\mathbf{b}_n]$, and $\Phi=[\phi_{y_1},\ldots,\phi_{y_n}]$. This is a quadratic function in $C$ and therefore a closed-form solution exists:
\begin{equation}
C = \frac{1}{\alpha + \gamma} (\alpha Z + \gamma \Phi)B^T(BB^T)^{-1}
\end{equation}

It is easy to observe that the optimization problem decomposes over each of the $p$ dimensions. Thus, we can reduce the computational cost by solving $p$ least square problem each with $mh$ variables.

\begin{equation}
\begin{gathered}
    \min_{C^{(t)}} \alpha\|Z^{(t)} - C^{(t)}B^{(t)}\|_2^2 + \gamma\|\Phi^{(t)} - C^{(t)}B^{(t)}\|_2^2 \\ 
     \forall \quad t=1,\ldots,p
\end{gathered}
\end{equation}
Each of the $p$ problems is a least squares problem with a closed form solution. Online learning algorithms can also be leveraged for acceleration~\cite{mairal2009online}.



\textbf{Updating $B$}. Given $\theta$, $\phi_{y_i}$ and $C$ fixed, optimizing binary matrix $B$, known as encoding phase, has been historically identified as the bottleneck of MCQ~\cite{babenko2014additive,martinez2016revisiting}. 

It can be seen that the composition indicator vector $\mathbf{b}_i$ is independent of all other vectors~$\{\mathbf{b}_t\}_{t \neq i}$. Thus, the optimization problem with respect to $B$ can be decomposed into $n$ independent subproblems:
\begin{equation}
\label{eq::optimizing_B}
\begin{aligned}
    &\min_{\mathbf{b}_i} & &\alpha\|\mathbf{z}_i - C\mathbf{b}_i\|_2^2 + \gamma\|\phi_{y_i} - C\mathbf{b}_i\|_2^2 \\
    & & & \mathbf{b}_i = [\mathbf{b}_{i1}^T,\ldots,\mathbf{b}_{im}^T]^T \\
    & & & \mathbf{b}_{ij} \in \{0,1\}^h, \|\mathbf{b}_{ij}\|_1=1 \\ 
    & & & i=1,\ldots,n\quad j=1,\ldots,m \\
\end{aligned}
\end{equation}

The problem is essentially a high-order Markov Random Field (MRF) problem which is NP-hard. Following~\cite{martinez2016revisiting}, we use Stochastic Local Search (SLS) method to optimize $\mathbf{b}_i$. The idea of SLS for escaping local minima is to iteratively alternate between a local search procedure, and a randomized pertubation to the current solution. For the local search, we again use alternative optimization technique. Given $\{\mathbf{b}_{ij}\}_{j\neq t}$ fixed, $\mathbf{b}_{it}$ is updated by exhaustively checking all codewords of $C_j$ and finding the element that minimizes the objective function in~(\ref{eq::optimizing_B}). For the perturbation procedure of SLS, we randomly choose $k$ codes by sampling from the uniform distribution $\mathcal{U}(1,m)$. The selected codes are perturbed by setting each of them to a uniformly selected random value between 1 and $h$. The resulting perturbed solution is then accepted as the starting point of the next local search procedure. Although this procedure is computationally demanding, it can be accelerated using GPU implementation~\cite{martinez2016solving,martinez2018lsq++}, making encoding even faster than codebook learning.


\subsection{Asymmetric Distance Computation}
Given the query, the search process starts by embedding the query using the trained network, $\mathbf{z}_{q} = f(\mathbf{q};\theta)$. Then, the inner product between $\mathbf{z}_{q}$ and all codewords are stored in $m\times h$ query-specific lookup table. Finally, inner product between query and all database vector is approximated with:

\begin{equation}
    \langle \mathbf{z}_{q}, \mathbf{z}_i\rangle \approx \sum_{j=1}^m \langle \mathbf{z}_{q}, C_j\mathbf{b}_{ij}\rangle
\end{equation}
Therefore, computing the inner product between query and each database item takes $O(m)$ lookups and $O(m)$ addition operations (same as PQ), plus the time required to embed the query into the deep feature space.

\subsection{Sparse Codebook Learning} In sparse codebook learning, the optimization problem is augmented with sparsity constraint on the codewords. The key advantage of sparse coodebooks is that the distance between the query and every codeword can be computed efficiently using sparse vector manipulations. This is practically important as for large codebooks, with many codewords, the time required for online construction of lookup tables become non-negligible. Zhang \etal~\cite{zhang2015sparse} have shown that sparse codewords can increase the search speed up to 30\%. As the name suggests, the Sparse Composite Quantization (SCQ) technique proposed in~\cite{zhang2015sparse} adds sparsity constraint to the CQ~\cite{zhang2014composite} formulation and uses coordinate descent to solve the optimization problem. However, CQ itself involves a hard optimization problem and adding the sparsity constraint makes the problem even harder.

In contrast, in our formulation, codebook optimization reduces to a linear regression problem, thus adding the sparsity constraint changes the objective to a regularized quadratic problem. In particular, using straightforward algebraic manipulations~(\ref{eq::codebook_problem}) can be rewritten as:
\begin{equation}
\begin{split}
(\alpha+ \gamma)\|\frac{\alpha Z + \gamma \Phi}{\alpha+\gamma} - CB\|_2^2 & -   \frac{\|\alpha Z + \gamma\Phi\|_2^2}{\alpha + \gamma}  \\
 & + \alpha\|Z\|_2^2 + \gamma\|\Phi\|_2^2
\end{split}
\end{equation}
Since only the first term depends on $C$, we can write the objective function of sparse quantization as:

\begin{equation}
   \min_{C} \|\frac{\alpha Z + \gamma \Phi}{\alpha+\gamma} - CB\|_2^2 \quad  s.t. \quad \|C\|_0 \leq \epsilon
\end{equation}
The resulting optimization is non-convex because of $L_0$ regularization term. Commonly, such problems are relaxed by replacing $L_0$ norm with convex $L_1$ norm. Therefore, our final objective function for learning sparse codebooks is defined as:
\begin{equation}
    \min_{C} \|\frac{\alpha Z^T + \gamma \Phi^T}{\alpha+\gamma} - B^TC^T\|_2^2 \quad  s.t. \quad  \|C\|_1 \leq \epsilon
\end{equation}
which is essentially a linear regression problem with $L_1$ norm regularization on the coefficients, known as Lasso in the statistical literature. It can be efficiently solved using a wide range of heavily-optimized off-the-shelf Lasso solvers such as feature-sign search~\cite{lee2007efficient} or SPGL1 solver~\cite{BergFriedlander:2008}.


\section{Experiments}
\label{sec::experiments}
In this section, we gauge the performance of the proposed supervised quantization approach by comparing it with the state-of-the-art against three different datasets..

\subsection{Datasets and Evaluation}
We conduct experiments on three standard datasets: CIFAR-10~\cite{krizhevsky2009learning}, NUS-WIDE~\cite{chua2009nus} and ImageNet~\cite{deng2009imagenet}.

CIFAR-10 dataset consists of 60,000 $32\times 32$ color images evenly divided into 10 categories. We follow the official split of the datasets and use 50K images as the training set and 10k images as the query set.

NUS-WIDE is a set of 269,648 images collected from Flickr. This is a multi-label dataset where each image is associated with one or multiple labels from a given 81 concepts. Following~\cite{shen2015supervised,wang2016supervised}, we collect 193,752 images that are from the 21 most frequent labels for evaluation, including \textit{sky, clouds,
person, water, animal, grass, building, window, plants, lake,
ocean, road, flowers, sunset, relocation, rocks, vehicles,
snow, tree, beach, and mountain}. For each label, we randomly sample 100 images as query points and the remaining images form the training set.

The dataset ILSVRC 2012, named as ImageNet in this paper, contains over 1.2 million images covering 1,000 categories. Following the settings in~\cite{cao2017hashnet,chena2018deep}, we select 100 categories and use images associated with them in the provided training set and the validation set as the training and the query sets, respectively.

\textbf{Parameter setting}. There are trade-off parameters in the objective function~(\ref{eq::final_optimization}): $\alpha$ for quantization loss, $\lambda$ for center loss and $\gamma$ for discriminative loss. We select parameters via validation. In particular, we choose a subset of the training set (same size as the query set), and the best parameters are chosen so that the average performance in terms of MAP is maximized against the validation set. We fix $\zeta$ to 0.5 and $k$ to 4. 

Following almost all MCQ techniques~\cite{babenko2014additive,norouzi2013cartesian,zhang2015sparse}, we choose $h=256$ to be the codebook size, so that each subindex fits into one byte of memory. This let us store $B$ as a $m \times n$ $\texttt{uint8}$ matrix. We vary $m=\{2,4,6,8\}$ such that $m\log_2h$ is equal to the desired bit-rates which are $\{16,32,48,64\}$. 

\textbf{Experimental settings.} We use the raw images as the input for all deep methods, but the images are resized to fit the input of the adopted model. For fairness of comparison, for all deep compact coding methods here, we use Alexnet as the core architecture. To reduce the size of deep features, we add a fully connected layer to the network which transforms the output of the network into a 256-dimensional feature space, thus $p=256$. We do not tune the size of feature space for saving time while we think that tuning it might yield better performance.
The $L_2$ normalization is performed on the 256-dimensional deep features using a $L_2$ normalize layer~\cite{ranjan2017l2}.

We fine-tune layers conv1–fc7 copied from the AlexNet
model pre-trained on ImageNet and train the last layer which maps the feature layer via
back-propagation.  As the last layer is trained from scratch, we set its learning rate to be 10 times that of the other layers. We use mini-batch stochastic gradient descent (SGD) with 0.9 momentum as the solver, and cross-validate the learning rate from $10^{-5}$ to $10^{-2}$ with a multiplicative step-size $\sqrt{10}$. We also fix the mini-batch size
of images as 128 and the weight decay parameter as 0.0005. Following~\cite{martinez2016revisiting}, we use SPGL1 as the lasso solver for the sparse extension of our algorithm~\cite{BergFriedlander:2008}. For non-deep methods, we extract the outputs of the layer ‘fc7’ in the deep model~\cite{donahue2014decaf} as input features.

\textbf{Methods}. We compare DSQ with a wide range of supervised compact coding methods including binary hashing methods: KSH~\cite{liu2012supervised}, ITQ~\cite{gong2013iterative}, SDH~\cite{shen2015supervised}, CNNH~\cite{xia2014supervised}, DPSH~\cite{li2016feature}, 
 DSH~\cite{liu2016deep}, HashNet~\cite{cao2017hashnet}, and supervised quantization techniques: SQ~\cite{wang2016supervised}, SUBIC~\cite{jain2017subic}, DQN~\cite{cao2016deep} and DTQ~\cite{liu2018deep}. We implemented SQ in Python as its source code is not available at the time of writing this paper. We tried our best to be faithful to the experimental settings of the paper~\cite{wang2016supervised}. Other techniques are executed using the implementation generously provided by the authors.

\subsection{Results}
\begin{table*}[]
    \centering
    \small
    \begin{tabular}{c c c c c c c c c c c c c}
        & \multicolumn{4}{c}{CIFAR-10} & \multicolumn{4}{c}{NUS-WIDE} & \multicolumn{4}{c}{ImageNet}  \\ \hline
         Method & 16 & 32 & 48 & 64 & 16 & 32 & 48 & 64 & 16 & 32 & 48 & 64 \\ \hline
         KSQ & 0.3216 & 0.3285 & 0.3371 & 0.3384 & 0.4061 & 0.4182 & 0.4264 & 0.4436 & 0.1620 & 0.2818 & 0.3422 & 0.3934 \\ 
         ITQ & 0.2412& 0.2432& 0.2482& 0.2531 & 0.5573 & 0.5932 & 0.6128 & 0.6166 & 0.3115 & 0.4632 & 0.5223 & 0.5446\\
         SDH & 0.4199 & 0.4301 & 0.4392 & 0.4465 & 0.5342 & 0.6282 & 0.6298 & 0.6335 & 0.2729 & 0.4521 & 0.5329 & 0.5893\\
         CNNH & 0.5373 & 0.5421  & 0.5765 & 0.5780 & 0.6221 & 0.6233 & 0.6321 & 0.6372 & 0.2888 & 0.4472 & 0.5328 & 0.5436\\
         DPSH & 0.6367 & 0.6412 & 0.6573 & 0.6676 & 0.7015 & 0.7126 & 0.7418 & 0.7423 & 0.3226 & 0.5436 & 0.6217 & 0.6534 \\ 
         DSH & 0.6192 & 0.6565  & 0.6624  & 0.6713  & 0.7181 & 0.7221 & 0.7521 & 0.7531 & 0.3428 & 0.5500 & 0.6329 & 0.6645\\ 
         HashNet & 0.6857 & 0.6923 & 0.7183 & 0.7187 & 0.7331 & 0.7551 & 0.7622 & 0.7762 & 0.5016 & 0.6219  & 0.6613 & 0.6824 \\
         DTQ & 0.7037 & 0.7191 & 0.7319 & 0.7373 & 0.7511 & 0.7812 & 0.7886 & 0.7892 & 0.5128  & 0.6123 & 0.6727 & 0.6916 \\
         SUBIC & 0.6555 & 0.6789& 0.6854 & 0.7014 & 0.7021 & 0.7131 & 0.7555 & 0.7568 & 0.5547 & 0.5597 & 0.6462 & 0.6622\\
         SQ & 0.6212 & 0.6438 & 0.6545 & 0.6578 & 0.7126 & 0.7138 & 0.7303 & 0.7423 & 0.3865 & 0.5586 & 0.6279 & 0.6618 \\
         DQN & 0.5979 & 0.6097 & 0.6099 & 0.6133 & 0.6913 & 0.7121 & 0.7471 & 0.7562 & 0.5065 & 0.6205 & 0.6669 & 0.6912\\
         DSQ & \textbf{0.7212} & \textbf{0.7346} & \textbf{0.7418} & \textbf{0.7589} & \textbf{0.7785} & \textbf{0.7899} & \textbf{0.7918} & \textbf{0.7988} & \textbf{0.5769} & \textbf{0.6541} & \textbf{0.6800} & \textbf{0.6940}\\ \hline
    \end{tabular}
    
    \caption{Single-domain category retrieval performance of DSQ versus the state-of-the-art with 16, 32, 48 and 64 bit codes.}
    \label{tab::single_domain}
\end{table*}


\textbf{Single domain retrieval}. Single-domain retrieval is the main experimental benchmark in the supervised binary hashing literature in which the query and training items belong to the same set of class labels. To evaluate performance of different techniques, we adopt the widely used Mean Average Precision (MAP). We report the results of MAP@5000 and MAP@1000 for NUS-WIDE and ImageNet datasets respectively. Table~\ref{tab::single_domain} shows the single-domain retrieval performance of DSQ against a wide-range of techniques. The observation is that our proposed method consistently delivers the best performance for different length of codes. We attribute the performance improvement to the proposed loss that aims at jointly preserving similarity information and controlling the quantization error. Also, dropping the orthogonality constraint increases the fidelity of codebooks which in turn reduces the approximation error of nearest neighbor search.
 Finally, back-proping the proposed supervised quantization loss can remarkably enhance the quantizibilty of the deep representation.

\begin{figure*}
 \centering
\subfloat{\input{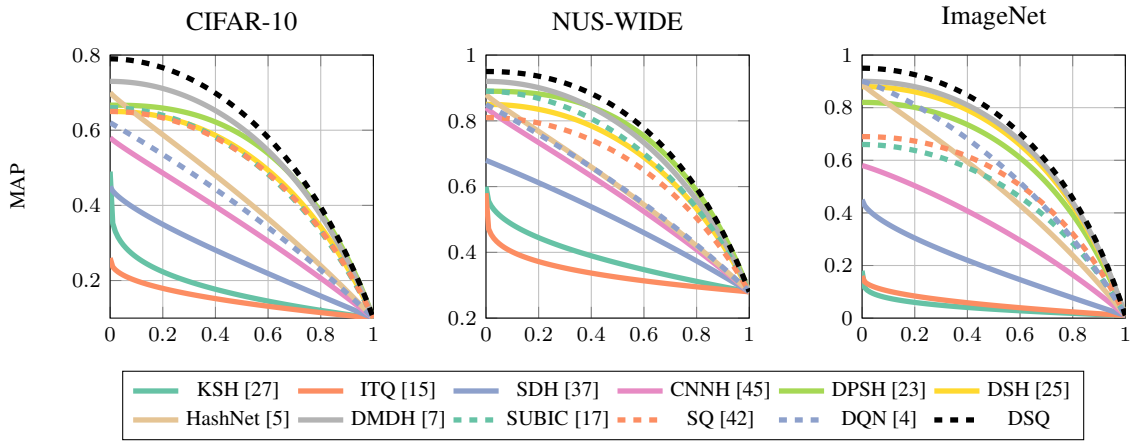}}
\caption{Precision-recall curves on the CIFAR-10, NUS-WIDE and ImageNet datasets for 64-bit codes.}
\label{fig::roc}
\end{figure*}

Figure~\ref{fig::roc} also shows the performance of different techniques in terms of the precision-recall curves for 64-bit codes. From the curves, we can observe that DSQ delivers higher precision than the state-of-the-art compact coding methods at the same recall rate. This shows that DSQ is also favourable for precision-oriented retrieval systems. Although the query time comparison is not presented here due to space limit, we observed that all deep MCQ techniques in this study exhibit similar query time mainly because they adopt the same core architecture (AlexNet). However, binary hashing techniques are often faster than deep MCQ as they incorporate Hamming distance to compare binary codes.

\textbf{Sparse coding.} We also show the performance of sparse extension of DSQ. To the best of our knowledge, sparse DSQ is the first attempt to explore supervised sparse multi-codebook quantization for semantic similarity search. Nevertheless, we compare our technique with two unsupervised sparse quantization techniques, SCQ~\cite{zhang2015sparse} and SLSQ~\cite{martinez2016revisiting} applied to the deep features of the ‘fc7’ layer of the deep model in~\cite{donahue2014decaf}.

Following~\cite{zhang2014composite}, we evaluate the sparse version of our algorithm using two degrees of sparsity: SDSQ1 with $\|C\|_0 \leq \epsilon = h\cdot p$ and SDSQ2 with $\|C\|_0 \leq \epsilon = h \cdot p+p^2$. Since the former criterion imposes a harder sparsity constraint on the codebooks, we would naturally expect to achieve lower search accuracy but better query time. We compare against SCQ1 and SCQ2 from~\cite{zhang2015sparse} and SLSQ1 and SLSQ2 from~\cite{martinez2016revisiting}. 

\begin{figure*}
\centering
\subfloat{\begin{tikzpicture}

\begin{axis}[%
width=3.5cm,
height=3.5cm,
scale only axis,
xmin=16,
xmax=64,
ymin=0.5,
ymax=0.74,
title={CIFAR-10},
ylabel={MAP},
label style={font=\footnotesize},
                    tick label style={font=\footnotesize}, 
                    legend style={nodes={scale=0.8, transform shape}},
axis background/.style={fill=white},
xmajorgrids,
ymajorgrids,
xtick={16,32,48,64},
legend style={legend cell align=left, align=left,font=\fontsize{7}{5}\selectfont,legend columns=2},
legend pos=south west
]
\addplot+[dashed,line width=1.5pt]
  table[row sep=crcr]{%
16	0.6912\\
32	0.7046\\
48	0.7097\\
64	0.7146\\
};
\addlegendentry{SDSQ1}

\addplot+[line width=1.5pt,dashed]
  table[row sep=crcr]{%
16	0.7098\\
32	0.7201\\
48	0.7218\\
64	0.7288\\
};
\addlegendentry{SDSQ2}

\addplot+[line width=1.5pt]
  table[row sep=crcr]{%
16	0.5779\\
32	0.6067\\
48	0.6099\\
64	0.6133\\
};
\addlegendentry{SCQ1~\cite{zhang2015sparse}}

\addplot+[line width=1.5pt,mark=diamond]
  table[row sep=crcr]{%
16	0.6079\\
32	0.608\\
48	0.6138\\
64	0.6133\\
};
\addlegendentry{SCQ2~\cite{zhang2015sparse}}

\addplot+[line width=1.5pt]
  table[row sep=crcr]{%
16	0.6279\\
32	0.6318\\
48	0.6425\\
64	0.6489\\
};
\addlegendentry{SLSQ1~\cite{martinez2016revisiting}}

\addplot+[line width=1.5pt]
  table[row sep=crcr]{%
16	0.6379\\
32	0.6518\\
48	0.6725\\
64	0.6789\\
};
\addlegendentry{SLSQ2~\cite{martinez2016revisiting}}

\end{axis}
\end{tikzpicture}%
}
\subfloat{
%
%

%
\begin{tikzpicture}

\begin{axis}[%
width=3.5cm,
height=3.5cm,
scale only axis,
xmin=16,
xmax=64,
ymin=0.66,
ymax=0.75,
title={NUS-WIDE},
label style={font=\footnotesize},
                    tick label style={font=\footnotesize}, 
                    legend style={nodes={scale=0.8, transform shape}},
axis background/.style={fill=white},
xmajorgrids,
ymajorgrids,
xtick={16,32,48,64},
legend style={legend cell align=left, align=left, draw=white!15!black}
]
\addplot+[dashed,line width=2.0pt]
  table[row sep=crcr]{%
16	0.7066\\
32	0.7166\\
48	0.7188\\
64	0.7333\\
};

\addplot+[line width=1.5pt,dashed]
  table[row sep=crcr]{%
16	0.7098\\
32	0.7201\\
48	0.7218\\
64	0.7488\\
};

\addplot+[line width=1.5pt]
  table[row sep=crcr]{%
16	0.6612\\
32	0.6865\\
48	0.6933\\
64	0.7036\\
};
\%addlegendentry{SCQ1}

\addplot+[line width=1.5pt]
  table[row sep=crcr]{%
16	0.6712\\
32	0.6899\\
48	0.7031\\
64	0.71\\
};

\addplot+[line width=1.5pt]
  table[row sep=crcr]{%
16	0.6799\\
32	0.6812\\
48	0.6999\\
64	0.7132\\
};

\addplot+[line width=1.5pt]
  table[row sep=crcr]{%
16	0.6899\\
32	0.6945\\
48	0.7011\\
64	0.7155\\
};

\end{axis}
\end{tikzpicture}%
}
\subfloat{
%
%

%
\begin{tikzpicture}

\begin{axis}[%
width=3.5cm,
height=3.5cm,
scale only axis,
xmin=16,
xmax=64,
ymin=0.52,
ymax=0.7,
title={ImageNet},
label style={font=\footnotesize},
                    tick label style={font=\footnotesize}, 
                    legend style={nodes={scale=0.8, transform shape}},
axis background/.style={fill=white},
xmajorgrids,
ymajorgrids,
xtick={16,32,48,64},
legend style={legend cell align=left, align=left, draw=white!15!black}
]
\addplot+[dashed,line width=2.0pt]
  table[row sep=crcr]{%
16	0.6123\\
32	0.6322\\
48	0.6602\\
64	0.675\\
};

\addplot+[line width=1.5pt,dashed]
  table[row sep=crcr]{%
16	0.6233\\
32	0.6422\\
48	0.6652\\
64	0.685\\
};

\addplot+[line width=1.5pt]
  table[row sep=crcr]{%
16	0.5212\\
32	0.5612\\
48	0.5812\\
64	0.5971\\
};

\addplot+[line width=1.5pt]
  table[row sep=crcr]{%
16	0.5312\\
32	0.5812\\
48	0.6012\\
64	0.6112\\
};

\addplot+[line width=1.5pt]
  table[row sep=crcr]{%
16	0.5312\\
32	0.5619\\
48	0.6\\
64	0.615\\
};

\addplot+[line width=1.5pt]
  table[row sep=crcr]{%
16	0.5512\\
32	0.5712\\
48	0.6112\\
64	0.6222\\
};

\end{axis}
\end{tikzpicture}%
}
\caption{Mean Average Precision performance of different sparse quantization techniques against three datasets.}
\label{fig::sparse_map}
\end{figure*}
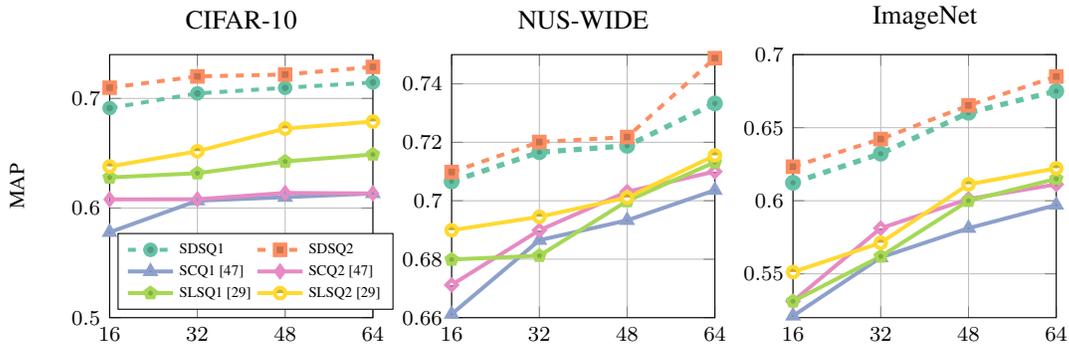

Figure~\ref{fig::sparse_map} shows the performance of different techniques against three different datasets. Again, in this scenario, we observe that sparse DSQ comfortably outperforms the baselines with a large margin mainly because sparse DSQ jointly optimizes the quantization error while preserving the semantic similarity and satisfying the sparsity constraint, whereas the other benchmarks separately apply unsupervised sparse quantization, which merely minimizes the quantization error.

\textbf{Cross-domain retrieval.} To further evaluate our supervised quantization method, we follow an alternative evaluation protocol from~\cite{sablayrolles2017should} wherein the model learned on a given set of training classes is tested on a new, disjoint set of test classes. This protocol is used to show how each method is capable of preserving the semantic information of certain classes implicitly even if the class samples are not included in the training set.

Toward this aim, we partition the samples based on their class labels such that 70\% of the labels belong to the training and the remaining labels are used to form the base and query set. Note that in this scenario the training set is used to optimize the parameters of the model. Once learning is completed, the training set is removed and the items of base set are mapped into compact codes using the trained model. Finally, the average performance over the query set is reported. We use 80\% of the samples with unseen classes as the training set and the rest as the query set. This process is repeated 5 times with random class splits and the average results is reported.
For this setting, during the encoding phase, we drop the $L_C$ term from loss because the trained centers do not correspond to any of the labels in the base set. Similarly, the regression loss term in SQ~\cite{wang2016supervised} is dropped during encoding as it directly depends on the class labels of training set. 

\begin{table}
\centering
\small
\begin{tabular}{c c c c c}
Method & 16 & 32 & 48 & 64 \\ \hline
CNNH & 0.6241 & 0.6456 & 0.6478 & 0.6491 \\
DPSH & 0.6894 & 0.7134 & 0.7198 & 0.7256 \\
HashNet & 0.7826 & 0.7941 & 0.7989 & 0.8010 \\ 
SUBIC & 0.7832 & 0.7931 & 0.8032 & 0.8077 \\ 
DSH & 0.7316 & 0.7388 & 0.7437 & 0.7456 \\ 
SQ & 0.7112 & 0.7126 & 0.7319 & 0.7389 \\ 
DQN & 0.7562 & 0.7612 & 0.7649 & 0.7655 \\
DTQ &  0.7525 & 0.7685 & 0.7700 & 0.7895 \\
DSQ &  \textbf{0.7944} & \textbf{0.8165} & \textbf{0.8195} & \textbf{0.8218} \\ \hline
\end{tabular}
\caption{Mean Average Precision performance of different techniques for the task of cross domain performance on CIFAR-10.}
\label{tab::cross_domain}
\end{table}

Table~\ref{tab::cross_domain} demonstrates the results of this experiment which shows the superiority of DSQ for different lengths of code. We also observe that the MAP performance of methods are generally higher than that of the previous protocol since there is less variation in the base set consisting of only 3 classes and fewer samples to retrieve from. Also the rank of techniques is different from the single domain experiments. For example, SUBIC exhibits the closest performance to DSQ whereas in single-domain setting DTQ is the closest.

\subsection{Ablation Study}
\begin{figure}
\centering
\begin{tikzpicture}
\begin{axis}[
	width=8cm,
	height=5cm,
    symbolic x coords={$L_{softmax}+L_Q$,$L_{softmax}+ L_Q + L_C$,$L_{softmax} + L_Q + L_D$,
    $L_C + L_D$,
    $L_{total}$,
    },
    label style={font=\small},
    tick label style={font=\small}, 
    xmajorgrids,
	ymajorgrids,
	ylabel={MAP},
    x tick label style={rotate=45, anchor=north east, inner sep=0mm}
    ]
    
    \addplot+[ybar,fill,mark=none, postaction={
        pattern=north east lines
    }] coordinates {
        ($L_{softmax}+L_Q$,0.54)
    };
    \addplot+[ybar,fill,mark=none, postaction={
        pattern=north east lines
    }] coordinates {
        ($L_{softmax}+ L_Q + L_C$,0.69)
    };
    \addplot+[ybar,fill,mark=none, postaction={
        pattern=north east lines
    }] coordinates {
    ($L_{softmax} + L_Q + L_D$,0.65)
     };
      \addplot+[ybar,fill,mark=none, postaction={
        pattern=north east lines
    }] coordinates {
    ($L_C + L_D$,0.15)
     };
     \addplot+[ybar,fill,mark=none, postaction={
        pattern=north east lines
    }] coordinates {
   ($L_{total}$,0.7589)
     };
\end{axis}
\end{tikzpicture}
\caption{Difference in MAP, when different loss components are excluded from DSQ objective function. The experiments are conducted on 64-bit codes of CIFAR-10 dataset.}
\label{fig::ablation}	
\end{figure}
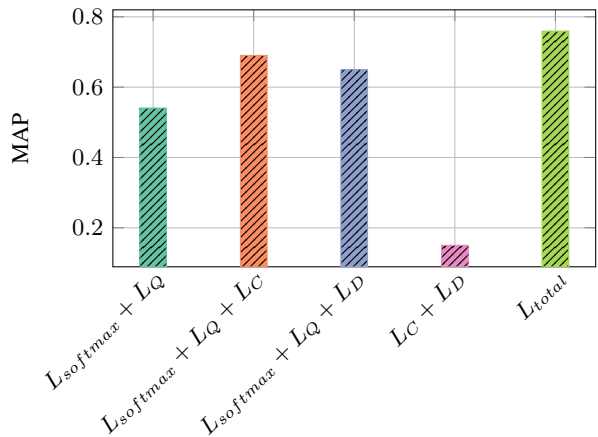

We also perform an ablation study to showcase the contribution and importance of loss function components on the final performance of the model by empirically comparing different variants of DSQ. We evaluate this experiment across different models to understand the sensitivity of DSQ to different terms: 1) $L_{softmax}+L_Q$, 2) $L_{softmax}+ L_Q + L_C$, 3) $L_{softmax} + L_Q + L_D$, and 4) $L_C + L_D$. For each model, the coefficients of different terms are again tuned using cross validation and the average performance of model for 64-bit codes against CIFAR-10 dataset is reported in Figure~\ref{fig::ablation}. 

The first observation is that all of the loss components contribute in improving MAP. Also, the plot indicates the importance of softmax loss. This is due to the fact that the softmax loss is the only term in the objective function that uses that class labels to force the deep features
of different classes staying apart, without it, the resulting loss function degrades all inputs points to be projected onto a single point. The figure also demonstrates considerable contribution of discriminative loss, $L_D$, showing the effectiveness of our framework in incorporating semantic information during quantization.

\section{Conclusion}
In this paper,  we propose a deep supervised quantization technique for efficient and fast image retrieval. By incorporating $L_2$ normalized features, we propose a simple yet efficient supervised MCQ algorithm for encoding unit normalized data points with similarity preserving binary codes. We also show that our algorithm can be easily extended to accommodate sparsity constraint in the codebooks which is necessary for learning large-scale codebooks. Comprehensive experiments justify that DSQ and its sparse extension generate compact binary codes that yield state-of-the-art retrieval performance on three standard benchmarks, namely CIFAR-10, NUS-WIDE, and ImageNet.

{\small
\bibliographystyle{ieee}
\bibliography{egbib}

\begin{thebibliography}{10}\itemsep=-1pt

\bibitem{babenko2014additive}
A.~Babenko and V.~Lempitsky.
\newblock Additive quantization for extreme vector compression.
\newblock In {\em CVPR}, pages 931--938, 2014.

\bibitem{cao2018hashgan}
Y.~Cao, B.~Liu, M.~Long, J.~Wang, and M.~KLiss.
\newblock Hashgan: Deep learning to hash with pair conditional wasserstein gan.
\newblock In {\em CVPR}, pages 1287--1296, 2018.

\bibitem{Cao_2018_CVPR}
Y.~Cao, M.~Long, B.~Liu, and J.~Wang.
\newblock Deep cauchy hashing for hamming space retrieval.
\newblock In {\em CVPR}, 2018.

\bibitem{cao2016deep}
Y.~Cao, M.~Long, J.~Wang, H.~Zhu, and Q.~Wen.
\newblock Deep quantization network for efficient image retrieval.
\newblock In {\em AAAI}, pages 3457--3463, 2016.

\bibitem{cao2017hashnet}
Z.~Cao, M.~Long, J.~Wang, and S.~Y. Philip.
\newblock Hashnet: Deep learning to hash by continuation.
\newblock In {\em ICCV}, pages 5609--5618, 2017.

\bibitem{Chen_2018_CVPR}
Z.~Chen, X.~Yuan, J.~Lu, Q.~Tian, and J.~Zhou.
\newblock Deep hashing via discrepancy minimization.
\newblock In {\em CVPR}, 2018.

\bibitem{chena2018deep}
Z.~Chena, X.~Yuana, J.~Lua, Q.~Tiand, and J.~Zhoua.
\newblock Deep hashing via discrepancy minimization.
\newblock In {\em CVPR}, pages 6838--6847, 2018.

\bibitem{chua2009nus}
T.-S. Chua, J.~Tang, R.~Hong, H.~Li, Z.~Luo, and Y.~Zheng.
\newblock Nus-wide: a real-world web image database from national university of
  singapore.
\newblock In {\em CIVR}, page~48, 2009.

\bibitem{deng2009imagenet}
J.~Deng, W.~Dong, R.~Socher, L.-J. Li, K.~Li, and L.~Fei-Fei.
\newblock Imagenet: A large-scale hierarchical image database.
\newblock In {\em CVPR}, pages 248--255, 2009.

\bibitem{donahue2014decaf}
J.~Donahue, Y.~Jia, O.~Vinyals, J.~Hoffman, N.~Zhang, E.~Tzeng, and T.~Darrell.
\newblock Decaf: A deep convolutional activation feature for generic visual
  recognition.
\newblock In {\em ICML}, pages 647--655, 2014.

\bibitem{eghbali2017online}
S.~Eghbali, H.~Ashtiani, and L.~Tahvildari.
\newblock Online nearest neighbor search in binary space.
\newblock In {\em ICDM}, pages 853--858, 2017.

\bibitem{eghbali2019online}
S.~Eghbali, H.~Ashtiani, and L.~Tahvildari.
\newblock Online nearest neighbor search using hamming weight trees.
\newblock {\em IEEE Trans. PAMI}, 2019.

\bibitem{eghbali2019fast}
S.~Eghbali and L.~Tahvildari.
\newblock Fast cosine similarity search in binary space with angular
  multi-index hashing.
\newblock {\em IEEE Trans. on Knowledge and Data Engineering}, 31(2):329--342,
  2019.

\bibitem{ge2014optimized}
T.~Ge, K.~He, Q.~Ke, and J.~Sun.
\newblock Optimized product quantization.
\newblock {\em IEEE Trans. PAMI}, 36(4):744--755, 2014.

\bibitem{gong2013iterative}
Y.~Gong, S.~Lazebnik, A.~Gordo, and F.~Perronnin.
\newblock Iterative quantization: A procrustean approach to learning binary
  codes for large-scale image retrieval.
\newblock {\em IEEE Trans. PAMI}, 35(12):2916--2929, 2013.

\bibitem{he2018triplet}
X.~He, Y.~Zhou, Z.~Zhou, S.~Bai, and X.~Bai.
\newblock Triplet-center loss for multi-view 3d object retrieval.
\newblock {\em arXiv preprint arXiv:1803.06189}, 2018.

\bibitem{jain2017subic}
H.~Jain, J.~Zepeda, P.~Perez, and R.~Gribonval.
\newblock Subic: A supervised, structured binary code for image search.
\newblock In {\em ICCV}, Oct 2017.

\bibitem{jegou2011product}
H.~Jegou, M.~Douze, and C.~Schmid.
\newblock Product quantization for nearest neighbor search.
\newblock {\em IEEE Trans. PAMI}, 33(1):117--128, 2011.

\bibitem{krizhevsky2009learning}
A.~Krizhevsky and G.~Hinton.
\newblock Learning multiple layers of features from tiny images.
\newblock Technical report, Citeseer, 2009.

\bibitem{kulis2009learning}
B.~Kulis and T.~Darrell.
\newblock Learning to hash with binary reconstructive embeddings.
\newblock In {\em NIPS}, pages 1042--1050, 2009.

\bibitem{lee2007efficient}
H.~Lee, A.~Battle, R.~Raina, and A.~Y. Ng.
\newblock Efficient sparse coding algorithms.
\newblock In {\em NIPS}, pages 801--808, 2007.

\bibitem{li2017deep}
Q.~Li, Z.~Sun, R.~He, and T.~Tan.
\newblock Deep supervised discrete hashing.
\newblock In {\em NIPS}, pages 2482--2491, 2017.

\bibitem{li2016feature}
W.-J. Li, S.~Wang, and W.-C. Kang.
\newblock Feature learning based deep supervised hashing with pairwise labels.
\newblock In {\em IJCAI}, pages 1711--1717, 2016.

\bibitem{liu2018deep}
B.~Liu, Y.~Cao, M.~Long, J.~Wang, and J.~Wang.
\newblock Deep triplet quantization.
\newblock {\em MM}, 2018.

\bibitem{liu2016deep}
H.~Liu, R.~Wang, S.~Shan, and X.~Chen.
\newblock Deep supervised hashing for fast image retrieval.
\newblock In {\em CVPR}, pages 2064--2072, 2016.

\bibitem{liu2014discrete}
W.~Liu, C.~Mu, S.~Kumar, and S.-F. Chang.
\newblock Discrete graph hashing.
\newblock In {\em NIPS}, pages 3419--3427, 2014.

\bibitem{liu2012supervised}
W.~Liu, J.~Wang, R.~Ji, Y.-G. Jiang, and S.-F. Chang.
\newblock Supervised hashing with kernels.
\newblock In {\em CVPR}, pages 2074--2081, 2012.

\bibitem{mairal2009online}
J.~Mairal, F.~Bach, J.~Ponce, and G.~Sapiro.
\newblock Online dictionary learning for sparse coding.
\newblock In {\em ICML}, pages 689--696, 2009.

\bibitem{martinez2016revisiting}
J.~Martinez, J.~Clement, H.~H. Hoos, and J.~J. Little.
\newblock Revisiting additive quantization.
\newblock In {\em ECCV}, pages 137--153, 2016.

\bibitem{martinez2016solving}
J.~Martinez, H.~H. Hoos, and J.~J. Little.
\newblock Solving multi-codebook quantization in the gpu.
\newblock In {\em ECCV}, pages 638--650, 2016.

\bibitem{martinez2018lsq++}
J.~Martinez, S.~Zakhmi, H.~H. Hoos, and J.~J. Little.
\newblock Lsq++: Lower running time and higher recall in multi-codebook
  quantization.
\newblock In {\em ECCV}, pages 491--506, 2018.

\bibitem{norouzi2013cartesian}
M.~Norouzi and D.~J. Fleet.
\newblock Cartesian k-means.
\newblock In {\em CVPR}, pages 3017--3024, 2013.

\bibitem{norouzi2012hamming}
M.~Norouzi, D.~J. Fleet, and R.~R. Salakhutdinov.
\newblock Hamming distance metric learning.
\newblock In {\em NIPS}, pages 1061--1069, 2012.

\bibitem{norouzi2014fast}
M.~Norouzi, A.~Punjani, and D.~J. Fleet.
\newblock Fast exact search in hamming space with multi-index hashing.
\newblock {\em IEEE Trans. PAMI}, 36(6):1107--1119, 2014.

\bibitem{ranjan2017l2}
R.~Ranjan, C.~D. Castillo, and R.~Chellappa.
\newblock L2-constrained softmax loss for discriminative face verification.
\newblock {\em arXiv preprint arXiv:1703.09507}, 2017.

\bibitem{sablayrolles2017should}
A.~Sablayrolles, M.~Douze, N.~Usunier, and H.~J{\'e}gou.
\newblock How should we evaluate supervised hashing?
\newblock In {\em ICASSP}, pages 1732--1736, 2017.

\bibitem{shen2015supervised}
F.~Shen, C.~Shen, W.~Liu, and H.~Tao~Shen.
\newblock Supervised discrete hashing.
\newblock In {\em CVPR}, pages 37--45, 2015.

\bibitem{BergFriedlander:2008}
E.~van~den Berg and M.~P. Friedlander.
\newblock Probing the pareto frontier for basis pursuit solutions.
\newblock {\em SIAM Journal on Scientific Computing}, 31(2):890--912, 2008.

\bibitem{Wang_2018_CVPR}
H.~Wang, Y.~Wang, Z.~Zhou, X.~Ji, D.~Gong, J.~Zhou, Z.~Li, and W.~Liu.
\newblock Cosface: Large margin cosine loss for deep face recognition.
\newblock In {\em CVPR}, 2018.

\bibitem{wang2018composite}
J.~Wang and T.~Zhang.
\newblock Composite quantization.
\newblock {\em IEEE Trans. PAMI}, 2018.

\bibitem{wang2018survey}
J.~Wang, T.~Zhang, N.~Sebe, H.~T. Shen, et~al.
\newblock A survey on learning to hash.
\newblock {\em IEEE Trans. PAMI}, 40(4):769--790, 2018.

\bibitem{wang2016supervised}
X.~Wang, T.~Zhang, G.-J. Qi, J.~Tang, and J.~Wang.
\newblock Supervised quantization for similarity search.
\newblock In {\em CVPR}, pages 2018--2026, 2016.

\bibitem{wen2016discriminative}
Y.~Wen, K.~Zhang, Z.~Li, and Y.~Qiao.
\newblock A discriminative feature learning approach for deep face recognition.
\newblock In {\em ECCV}, pages 499--515, 2016.

\bibitem{wu2017multiscale}
X.~Wu, R.~Guo, A.~T. Suresh, S.~Kumar, D.~N. Holtmann-Rice, D.~Simcha, and
  F.~Yu.
\newblock Multiscale quantization for fast similarity search.
\newblock In {\em NIPS}, pages 5745--5755, 2017.

\bibitem{xia2014supervised}
R.~Xia, Y.~Pan, H.~Lai, C.~Liu, and S.~Yan.
\newblock Supervised hashing for image retrieval via image representation
  learning.
\newblock In {\em AAAI}, page~2, 2014.

\bibitem{zhang2014composite}
T.~Zhang, C.~Du, and J.~Wang.
\newblock Composite quantization for approximate nearest neighbor search.
\newblock In {\em ICML}, pages 838--846, 2014.

\bibitem{zhang2015sparse}
T.~Zhang, G.-J. Qi, J.~Tang, and J.~Wang.
\newblock Sparse composite quantization.
\newblock In {\em CVPR}, pages 4548--4556, 2015.

\end{thebibliography}
}

\end{document}